\documentclass{article}

\usepackage[preprint,nonatbib]{nips_2018}

\usepackage[utf8]{inputenc} 
\usepackage[T1]{fontenc}    
\usepackage{hyperref}       
\usepackage{url}            
\usepackage{booktabs}       
\usepackage{amsfonts}       
\usepackage{nicefrac}       
\usepackage{microtype}      
\usepackage[pdftex]{graphicx} 
\usepackage{subcaption} 
\usepackage{amsmath} 
\usepackage{wrapfig} 
\usepackage{lipsum} 

\newcommand{\task}[1]{\emph{#1}}

\title{Towards continual learning in medical imaging}

\author{
  Chaitanya Baweja \\
  Imperial College London \\
  \texttt{cb5117@imperial.ac.uk} \\
  \And
  Ben Glocker \\
  Imperial College London \\
  \texttt{b.glocker@imperial.ac.uk} \\
  \And
  Konstantinos Kamnitsas \\
  Imperial College London \\
  \texttt{kk2412@imperial.ac.uk} \\
}

\begin{document}

\maketitle

\begin{abstract}
This work investigates continual learning of two segmentation tasks in brain MRI with neural networks.
To explore in this context the capabilities of current methods for countering catastrophic forgetting of the first task when a new one is learned, we investigate \emph{elastic weight consolidation} \cite{kirkpatrick2017overcoming}, a recently proposed method based on Fisher information, originally evaluated on reinforcement learning of Atari games. We use it to sequentially learn segmentation of normal brain structures and then segmentation of white matter lesions.
Our findings show this recent method reduces catastrophic forgetting, while large room for improvement exists in these challenging settings for continual learning.
\end{abstract}


\section{Introduction}

Advances in machine learning have led to rapid developments towards automation of various tasks, such as detection of pathology in medical scans. The currently most successful models are supervised neural networks. A network is trained using a manually annotated database for a specific task. However the amount of potential tasks in healthcare is immense. Thus annotating a large database for each task, to train specialized models from scratch, is not a scalable solution. Instead we envision that a model with pre-acquired knowledge about common tasks could be supplied to clinicians and, according to their needs, they could further train it to perform a new task. First, by utilizing existing knowledge from previous tasks, the model should quickly grasp the new task with only limited supervision. Furthermore, it should be able to incorporate the new knowledge with existing one, improving its knowledge both for the original and any future tasks. Finally, learning a new task should be possible without access to training data for the earlier tasks, which may no longer be available.

Such sequential knowledge acquisition is known as \emph{continual} and \emph{lifelong learning} \cite{thrun1998lifelong}. It is related to \emph{multitask learning} \cite{caruana1997multitask}, which assumes training data for all tasks of interest are available and all tasks are learnt concurrently. Instead, here, we assume that when learning a new task, the training data for previous tasks are no longer available. Sequential learning poses a great challenge for neural networks, known as \emph{catastrophic forgetting} \cite{mccloskey1989catastrophic,ratcliff1990connectionist,mcclelland1995there}, where knowledge about an old task is lost when changing a network's parameters during training to meet the objective for a new task.


Countering catastrophic forgetting in neural networks has recently attracted increased research attention. The first category of works derive regularization costs such that knowledge of the new task can be incorporated in existing capacity while preserving model behaviour on the old task \cite{li2017learning, kirkpatrick2017overcoming}. Other approaches extend a network with extra capacity or components for each new task \cite{rusu2016progressive, rebuffi2017learning}. This may alleviate forgetting but does not effectively fuse old and new knowledge. Such approach was applied for supervised domain adaptation of a model to different MRI scanners \cite{karani2018lifelong}, but not for different tasks, where the label spaces and labelling functions differ.

This work explores catastrophic forgetting when learning sequentially two different tasks in medical imaging: segmentation of normal structures and segmentation of white matter lesions in brain MRI. We investigate the potential of a recently proposed method, \emph{Elastic Weight Consolidation} (EWC) \cite{kirkpatrick2017overcoming}, originally evaluated for reinforcement learning of Atari games. The method tries to preserve network connections important for previous tasks, by regularizing connections with high Fisher information. We show experimentally that EWC reduces catastrophic forgetting in our settings. This study, the first of its kind in medical imaging, indicates that there is potential in this approach, while showing that there is significant space for further research and improvements towards continual learning.

\section{Fisher information for continual learning with neural networks}
\label{sec:method}

Suppose we have some data $\mathcal{D}_A$, such as images $x$ and corresponding annotations $y$, that reflect a particular task \emph{A}. A discriminative neural network parameterized by $\theta$ learns to approximate the distribution $p(y|x,\theta)$ that generates $y$ given $x$. For this, it learns the optimal parameters $\theta_{A}^{*}$ during training in order to minimize an appropriate loss $\mathcal{L}_A(\mathcal{D}_A,\theta)$. In continual learning, after training for task \emph{A}, it is assumed that $\mathcal{D}_A$ is no longer available. We are then given new training data, $\mathcal{D}_B$, for another task \emph{B}. Starting from the existing knowledge encapsulated in $\theta_{A}^{*}$, we wish to further change the parameters to also solve \emph{B}, while preserving the knowledge about \emph{A}. Assuming large, over-parameterized neural networks, many configurations of $\theta$ may lead to similar performance (\cite{hecht1992theory,choromanska2015loss}). Thus it is likely that there is a solution for task \task{B}, $\theta^*_B$ that is in the neighbourhood of $\theta^*_A$. Staying near $\theta^*_A$ during training on \task{B} can be encouraged by a regularizer based on L2 distance $R(\theta)=(\theta_i - \theta^*_{A,i})^2$, but this does not guarantee meaningful minima with respect to $A$.

Instead, one can investigate the importance of each parameter $\theta_i$ with respect to behaviour of the model. A measure for this is \emph{Fisher information}, which expresses the amount of information observing variable $Y$ carries about a parameter $\theta$ that models distribution of $Y$. For a model with $\theta \in \mathbb{R}^K$ parameters, this is expressed by the \emph{Fisher Information Matrix} $F$ defined as: 
\begin{equation}
F = \mathop{\mathbb{E}}_{(x,y)} \left[ \nabla_{\theta} \log p(y|x,\theta) \nabla_{\theta} \log p(y|x,\theta)^T \right]
\in \mathbb{R}^{K\!\times\!K}
\end{equation}
It is the variance of the \emph{score function} $s(\theta)\!=\!\nabla_{\theta} \log p(y|x,\theta)$, the expected value of which is zero.
$F$ quantifies how much a change of a parameter's value is expected to affect the output of a network $p(y|x,\theta)$. Intuitively, if $\nabla_{\theta_i} \log p(y|x,\theta_i)\!=\!0$, then $F_{i,j}\!=\!0,\ F_{j,i}\!=\!0,\ \forall j$, which expresses that parameter $\theta_i$ can be altered without change of the output. 
For continual learning, it is possible to use $F$ to regularize the change of each parameter when training for \task{B} according to its \emph{importance} for task \task{A} \cite{kirkpatrick2017overcoming}.
Because $K$ can be very large for neural networks, $F$ can be impractical to compute however.

\emph{Elastic Weight Consolidation} \cite{kirkpatrick2017overcoming} is a regularizer that is based on the assumption that the parameters of a network are uncorrelated (weak assumption if $K$ is very large). In this case $F$ is diagonal, thus one needs to compute only $K$ values. \emph{Importance} of parameter $\theta_i$ for task \task{A} is then:  
\begin{equation} \label{eq:fisher_i}
F_i = \mathop{\mathbb{E}}_{(x,y)\sim\mathcal{D}_A}\left[ \left( \nabla_{\theta_{A,i}^{*}} \log p(y|x,\theta_{A,i}^{*}) \right)^2 \right] \in \mathbb{R}
\end{equation}
which is computed after convergence of training on \task{A}. Finally, training for task \task{B} is performed by minimizing the follow total cost:
\begin{equation} \label{eq:loss_ewc_total}
    \mathcal{L}_{B,Total}(\mathcal{D}_B,\theta, \theta^{*}_{A}) = \mathcal{L}_{B}(\mathcal{D}_B,\theta) + \lambda \sum_{i=1}^{K} F_i (\theta_i - \theta^{*}_{A,i})^2
\end{equation}
EWC protects the parameters with high $F_i$ to stay close to the values needed for \task{A}, while parameters with low $F_i$ are allowed to move more freely, constituting capacity of the network allocated to task \task{B}. $\lambda$ controls strength of regularization. In what follows, we investigate the potential of EWC for the first time on a challenging biomedical application for sequential learning of two different tasks.

\section{Evaluation}

\subsection{Experimental setup}

\paragraph{Databases:}We use images and corresponding pixel-wise labels provided by UK Biobank \cite{miller2016multimodal} for the following tasks. \textbf{Task \task{A}}: multi-class segmentation of cerebrospinal fluid (CSF), grey matter (GM), white matter (WM). \textbf{Task \task{B}}: segmentation of white matter lesions (WML). We select 275 cases where WML is present. 87 cases are used when training for \task{A}, 88 when training for \task{B}, 100 to validate both tasks. In all experiments we use T1 and Flair sequences, after z-score normalization.

\paragraph{Experiments:}
We use the DeepMedic (DM) 3D convnet \cite{kamnitsas2017efficient} with default configuration, for its reliable performance in segmenting volumetric scans. We performed the following experiments:\\
\textit{DM-A}: Train DM only for task \task{A} from scratch.\\
\textit{DM-B}: Train DM only for task \task{B} from scratch.\\
\textit{Multi-task}: learn \task{A} and \task{B} jointly, using a DM with two classification layers. This is an upper bound for performance of continual learning, where data $\mathcal{D}_A$ is assumed unavailable when learning \task{B}.\\
\textit{Fine-tune}: Add new classification layer for task \task{B} on pre-trained DM-A, fine-tune whole net on $\mathcal{D}_B$.\\
\textit{L2}: Similar to \textit{fine-tune}, but regularize training for \task{B} via $L_2$ distance of $\theta$ from $\theta^*_{A}$ ($F_i\!=\!1 \forall i$ in Eq.~\ref{eq:loss_ewc_total}).\\
\textit{EWC}: Compute $F_i$ for each $\theta_i$ of pre-trained DM-A according to Eq.~\ref{eq:fisher_i}. Then add a new classification layer for task \task{B} on pre-trained DM-A, and learn task \task{B} by minimizing loss given in Eq.~\ref{eq:loss_ewc_total}. 

\subsection{Results}
\label{subsec:results}

\begin{figure}
\centering
\begin{minipage}[b]{0.30\textwidth}
\caption{Starting with DM pre-trained on task \task{A} (epoch 0), we train it for task \task{B} for 20 epochs, using (top) $L_2$ regularizer or (bottom) EWC, with varying $\lambda$ values. Plots show evolution of segmentation performance (DSC\%) for classes of task \task{A} (three left columns) and task \task{B} on random patches from validation cases.}
\label{fig:plots}
\end{minipage}
\hspace{0.03\textwidth}
\begin{subfigure}[b]{0.65\textwidth}
  \centering
  \includegraphics[width=\linewidth]{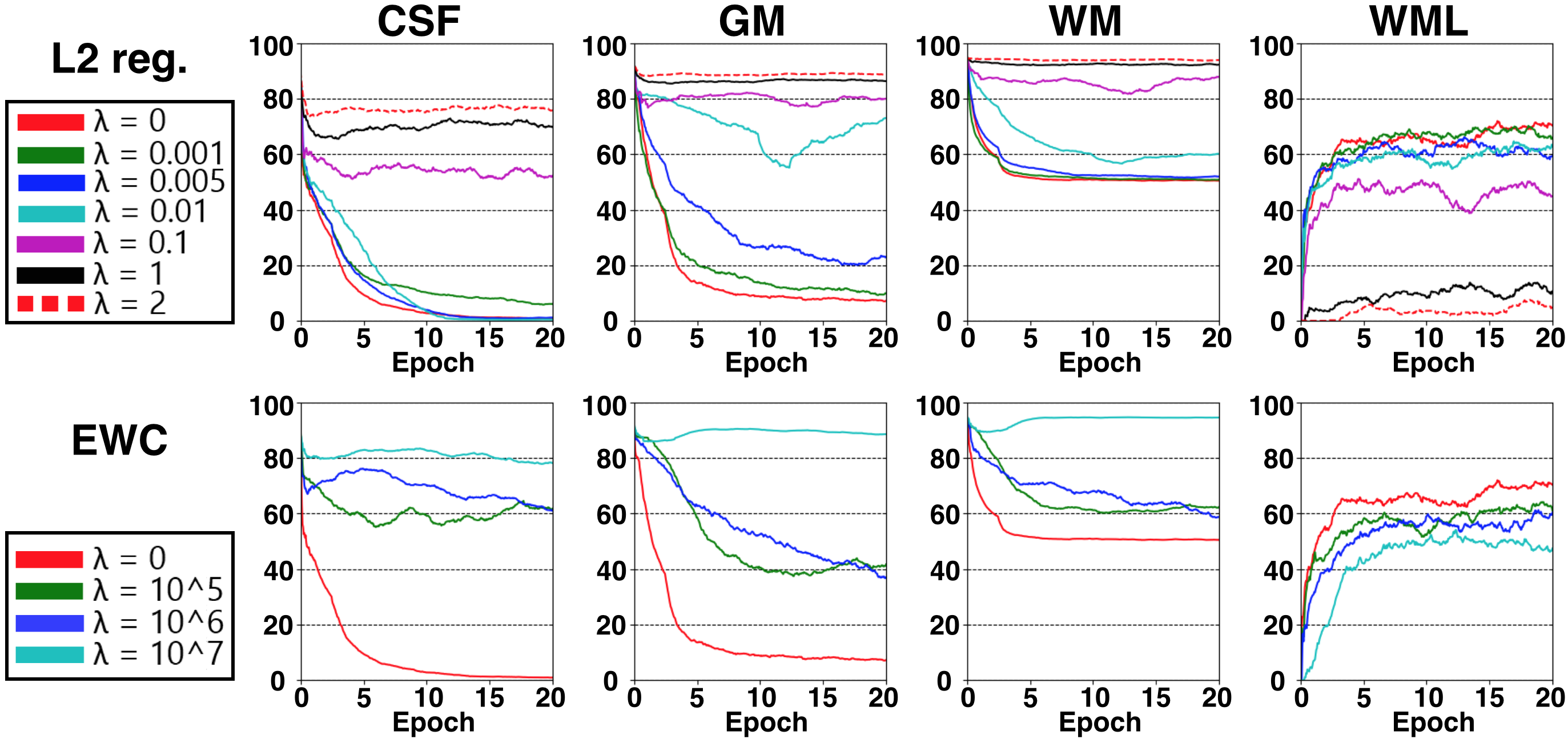}
\end{subfigure}
\end{figure}

\begin{wrapfigure}{r}{0.37\textwidth}
\vspace{-30pt}
\centering
\fontsize{6.8}{6.8}\selectfont
\caption{DSC\% achieved by employed methods when fully segmenting the validation cases.}
\label{tab:results}
\begingroup
\setlength{\tabcolsep}{5pt}
\begin{tabular}{lcccc}
\toprule
& \multicolumn{3}{c}{Task A}            & Task B    \\
\cmidrule(r){2-4} \cmidrule(r){5-5}
Method          & CSF & GM & WM         & WML \\
\midrule
DM-A            & 89.4 & 92.5 & 95.0    & -     \\
DM-B            & -     & -     & -     & 61.3  \\
Multi-task     & 88.9 & 92.2 & 94.8    & 63.8  \\
Fine-tune       & 00.9 & 11.9 & 50.8    & 62.1 \\
L2 $\lambda\!=\!0.005$ 
                & 00.5 & 07.9 & 50.7    & 54.4      \\
L2 $\lambda\!=\!0.01$ 
                & 02.9 & 80.6 & 59.9    & 56.3      \\
L2 $\lambda\!=\!0.1$ 
                & 50.8 & 82.0 & 87.5    & 42.3      \\
EWC $\lambda\!=\!10^6$ 
                & 60.1 & 32.6 & 62.5    & 53.4  \\
EWC $\lambda\!=\!10^7$ 
                & 79.2 & 88.7 & 94.4    & 44.5  \\
\bottomrule
\end{tabular}
\endgroup
\\[5ex] 
\begin{subfigure}[b]{0.37\textwidth}
  \centering
  \includegraphics[width=\linewidth]{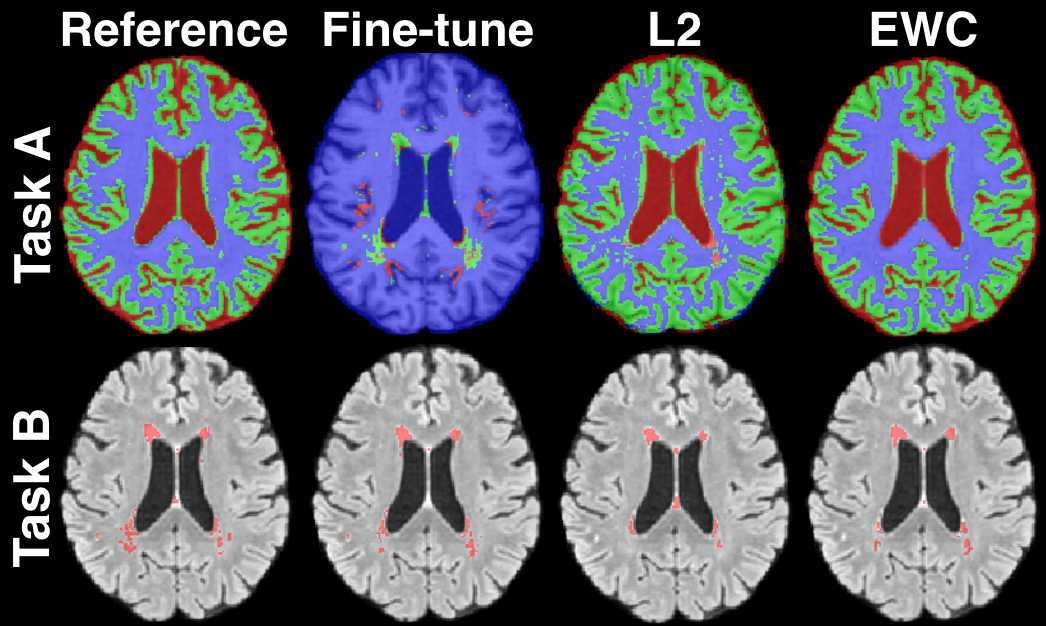}
\end{subfigure}
\caption{Segmentations from employed methods after learning task \task{B}. EWC mitigates forgetting of task \task{A}.}
\label{fig:visual_results}
\vspace{-35pt}
\end{wrapfigure}

We evaluate segmentation performance of the employed models on the validation subjects and report the Dice similarity coefficient (DSC) for each class in Table~\ref{tab:results}. Catastrophic forgetting is not an issue in multi-task learning, as data for both tasks are available during training. In contrast, unregularized \textit{fine-tuning} for task \task{B} suffers from it, completely forgetting task \task{A}. Regularizing training on task \task{B} via the naive $L_2$ regularizer or EWC mitigates forgetting. The strength of regularization $\lambda$ (Eq.~\ref{eq:loss_ewc_total}) acts as a trade-off between learning task \task{B} and forgetting task \task{A}. By comparing settings where $L_2$ and EWC regularization perform similarly on task \task{B}, we see that EWC preserves higher performance in task \task{A}, although there is still a gap until the upper bound set by multi-task learning. We show visual examples of the results for $L_2$ ($\lambda\!=\!0.1$) and EWC ($\lambda\!=\!10^{7}$) regularization in Fig.~\ref{fig:visual_results}. 

Finally, we present how segmentation performance on patches extracted randomly from validation subjects evolves while training for task \task{B} using $L_2$ and EWC regularizers with different values of $\lambda$ in Fig.~\ref{fig:plots}. The trade-off between learning task \task{B} and forgetting task \task{A} can be observed, while EWC performs overall better than the basic $L_2$ regularizer.

\section{Conclusion}
\label{sec:conclusion}

We investigated continual learning of two different segmentation tasks in the context of brain MRI. We explored the potential of EWC \cite{kirkpatrick2017overcoming}, a recently proposed regularizer based on Fisher information, previously shown promising for continual reinforcement learning on Atari games.
Our investigation on bio-medical data shows that the technique is promising for alleviating catastrophic forgetting. The findings of this work, one of the firsts of its kind on medical imaging data, also show that current methods leave significant space for further research before continual learning becomes practical. We hope this work will stimulate further investigations in this largely unexplored area.

\section*{Acknowledgments}

This project has received funding from the European Research Council (ERC) under the European Union's Horizon 2020 research and innovation programme (grant agreement No 757173, project MIRA, ERC-2017-STG). KK is supported by the President’s PhD Scholarship of Imperial College London. This research has been conducted using the UK Biobank Resource under Application Number 12579.

\small

\bibliographystyle{unsrt} 
\bibliography{refs}

\end{document}